\documentclass[10pt,letterpaper]{article}

\usepackage{cogsci}

\cogscifinalcopy % Uncomment this line for the final submission 

\usepackage{hyperref}
\usepackage{pslatex}
\usepackage{apacite}
\usepackage{graphicx}
\usepackage{multirow}
\usepackage{float} % Roger Levy added this and changed figure/table
\title{LLM Agents Display Human Biases but Exhibit Distinct Learning Patterns}
 
\author{{\large \bf Idan Horowitz (idanhorowitz@campus.technion.ac.il)} \\
  Faculty of Data and Decision Sciences, Technion \\
  Haifa, 3200003, Israel
  \AND {\large \bf Ori Plonsky (plonsky@technion.ac.il)} \\
 Faculty of Data and Decision Sciences, Technion \\
  Haifa, 3200003, Israel}

\begin{document}

\maketitle

\begin{abstract}
We investigate the choice patterns of Large Language Models (LLMs) in the context of Decisions from Experience tasks that involve repeated choice and learning from feedback, and compare their behavior to human participants. We find that on the aggregate, LLMs appear to display behavioral biases similar to humans: both exhibit underweighting rare events and correlation effects. However, more nuanced analyses of the choice patterns reveal that this happens for very different reasons. LLMs exhibit strong recency biases, unlike humans, who appear to respond in more sophisticated ways. While these different processes may lead to similar behavior on average, choice patterns contingent on recent events differ vastly between the two groups. Specifically, phenomena such as ``surprise triggers change" and the ``wavy recency effect of rare events" are robustly observed in humans, but entirely absent in LLMs. Our findings provide insights into the limitations of using LLMs to simulate and predict humans in learning environments and highlight the need for refined analyses of their behavior when investigating whether they replicate human decision making tendencies.

\textbf{Keywords:} language models, LLMs, decisions from experience, recency bias, decision making, surprise, learning
\end{abstract}

\section{Introduction}\label{intro}

Large Language Models (LLMs), particularly in the form of AI chatbots, assistants, or agents, have become a dominant focus over the last few years, with enormous potential to aid and improve global efficiency, while perhaps also presenting a threat in their incredibly human-seeming conversational abilities. As such, the volume of research around them and their possible applications has grown enormously. As probabilistic models, LLMs may not be susceptible to the same behavioral tendencies that humans commonly exhibit, making them difficult to understand, study and use in a predictable manner. Additionally, the rapid advancements in LLMs call for continuous re-evaluation of these findings. In this study, we evaluate LLMs in a domain previously unexplored in that respect: Decisions From Experience (DFE) tasks.

In DFE tasks, participants face a repeated choice between options that are each associated with some payoff distribution and get feedback on their choices. Usually, participants do not have a-priori knowledge of the the payoff distribution, and should thus learn it from the feedback. These studies allow exploration of the basic learning processes that emerge when people repeatedly engage with an environment and learn it from experience. Studies of DFE reveal several highly robust decision-making phenomena \cite{Erev2023}, and in this work, we investigate whether LLMs also exhibit similar behaviors. 

\subsection{Decisions from Experience Phenomena} Our focus is on four distinct behavioral phenomena that emerge in DFE studies. One of the most robust of these phenomena is \emph{underweighting of rare events} \cite{BarronErev2003, hertwig2004}. When making decisions based on experience with the choice task, people tend to behave as if they give rare events lower weight than they objectively deserve. Underweighting of rare events sharply contrasts the pattern predicted by Prospect Theory \cite{prospect} (that pertains to one-shot decisions based on a description of the task). The difference between the way people treat rare events when basing their decisions on experience vs. a description of the choice task is known as the ``description-experience gap'' \cite{hertwig2009}.

A second behavioral phenomenon that emerges in DFE studies is \emph{surprise triggers change} \cite{stc}, according to which a surprising event in one trial increases the chances that a decision maker will switch their selection in the next trial. This occurs even when the payoffs in that previous trial were reinforcing of their most recent choice.  

A third observation from DFE studies is the \emph{wavy recency effect} of rare events \cite{wavy-recency}, which refers to the sequential dependencies that people exhibit in consecutive trials. Individuals appear to respond to rare events in a wavy pattern: in the trials that immediately follow a rare event, they choose as if their sensitivity to the rare event (a) decreases for a few trials, usually reaching a trough 3 trials after the rare event, and then, somewhat oddly, (b) increases to a peak approximately 12-15 trials after the rare event (i.e., they behave as if the rare event becomes gradually \emph{more} likely to re-occur), and then diminishes back. Together, these changing sensitivities make a wave-like pattern of choice. 

Finally, when facing decision tasks in which the realized outcomes of the choice options are correlated, subjects exhibit a \emph{``correlation effect"} \cite{correlation-effect} suggesting that they quickly learn to choose the best option when the (usually positive) correlation implies that option consistently outperforms the alternatives, but learn slowly when the (usually negative) correlation implies that the best option (which is better on average) is not consistently better than the alternatives.

\subsection{Our Contribution}  
A thorough investigation of Decisions from Experience (DFE) studies, and particularly the phenomena mentioned above, has yet to be conducted on LLM agents. The goal of this research is to explore the behavior of LLM agents in common DFE tasks, and compare their decisions in these tasks to those of humans. We also explore how the manner by which the history of decisions is provided to the agent, and how the LLM's temperature, the model's degree of variability, influence the results, if at all. Finally, we try to infer the manner by which the LLM agents learn by comparing their behavior to several basic choice strategies. A better understanding of these issues has the potential to aid in the use of LLMs to generate human-like data and to improve on the current methods used in modeling and predicting human behavior. 

\section{Existing Work}
Numerous studies have attempted to assess the behavior of LLMs through the lens of cognitive psychology (e.g.\citeNP{Binz_assesment, codaforno2024cogbenchlargelanguagemodel, jones}), with results inconclusive. While language models display many human-like intuitions \cite{Hagendorff_2023} and behavioral similarities, including abstract reasoning \cite{yehudai2024nurse}, communication \cite{liu2024largelanguagemodelsnavigate, liu2023improvinginterpersonalcommunicationsimulating} and decision making patterns (including in the errors and biases that arise from such tasks)\cite{mei2023, rathje_identify_constructs, lampien_dasgupta}, there are also areas in which the irrationality of LLMs does not align with that of humans \cite{macmillanscott2024_irrationality}. They also tend to exhibit certain distortions \cite{aher2023, new-paper}, have low social intelligence \cite{wang2024evaluating} and perform poorly in social scenarios with information asymmetry \cite{zhou2024_information_assymetry}. For example, \citeauthor{nobandegani2024decision} \citeyear{nobandegani2024decision} found that while LLMs behave rationally in response to the Allais Paradox, they fail to do so for the Ellsberg Paradox. Nonetheless, the similarities to humans in the way LLMs behave have led to successes in using LLMs to recover findings in human experiments and therefore simulate human behavior \cite{aher2023, Argyle_2023_simulate_samples, coletta2024llmdrivenimitationsubrationalbehavior, horton2023largelanguagemodelssimulated, shapira1}. These works lay the foundation to integrate LLMs in the prediction of human decision-making \cite{aher2023, binz2023, shapira1}. None of the works above, however, focus on many-shot, repeated decisions. 

In the multi-shot domain, \citeauthor{zhu2024prob-estimation} \citeyear{zhu2024prob-estimation} found that LLMs estimate probabilities in a similar manner to humans, while \citeauthor{brown2020languagemodelsfewshotlearners} \citeyear{brown2020languagemodelsfewshotlearners} establish that LLMs are better learners in language tasks after a few shots, as opposed to zero- or one-shot learning. \citeauthor{rl-experience-llms} \citeyear{rl-experience-llms} compared the choices between humans and LLMs in repeated trials from a Reinforcement Learning perspective, but do not analyze cognitive patterns emerging from these tasks. Thus, no work exists covering phenomena arising from many repeated choices in LLMs.

\begin{table*}
\caption{Task parameters and \textit{B} choice rates (\%)} 
\label{tab_tasks} 
\vskip 0.12in
\centering
\resizebox{\textwidth}{!}{%
\begin{tabular}{ccccccccccccc}
\hline
\multicolumn{6}{c}{Task parameters} & & \multicolumn{5}{c}{Choice rate of option \textit{B} (\%)} \\
\cline{1-6} \cline{8-12} % Horizontal lines under each group
Task & A payoff distribution & B payoff distribution & $Corr(A, B)$ & $EV(A)$ & $EV(B)$ & & Gemini-all & Gemini-chat & GPT-all & GPT-chat & Humans \\
\cline{1-6} \cline{8-12} % Horizontal lines under each group
\textbf{1} & 0 $w.p$ 1; else 0 & 1 $w.p$ 0.9; else -10 & 0 & 0 & -0.1 & & 74 & 85 & 90 & 82 & 71  \\
\textbf{2} & 0 $w.p$ 1; else 0 & -1 $w.p$ 0.9; else 10 & 0 & 0 & 0.1 & & 16 & 14 & 33 & 49 & 28  \\
\textbf{3} & 6 $w.p$ 0.5; else 0 & 8 $w.p$ 0.5; else 0 & 1 & 3 & 4 & & 96 & 67 & 77 & 59 & 91 \\
\textbf{4} & 6 $w.p$ 0.5; else 0 & 9 $w.p$ 0.5; else 0 & -1 & 3 & 4.5 & & 61 & 50 & 53 & 51 & 65  \\
\hline
\end{tabular}}
\vspace{0.1cm} % Adds a bit of space between the table and the note
\footnotesize
\textit{Note:} The notation ``$x$  $w.p$  $p$; else $y$" refers to ``payoff of $x$ with probability $p$ and payoff of $y$ otherwise". $Corr(A, B)$ measures the linear correlation between the payoffs of $A$ and $B$. $EV(x)$ is the expected value for option $x$.
\end{table*}

\section{Methods}

\subsection{Experimental Design}\label{decs}
We presented both LLM agents and human participants with one of four binary-choice DFE tasks. The tasks' design is similar to those commonly used in previous DFE experiments (e.g. \citeNP{Erev2017, Erev2023, trial-citations}). Each task involved repeated choice between the same two options, \textit{A} or \textit{B}, for 100 trials. Decision makers were not given information regarding the payoff distributions of either choice option, but could learn it from feedback: After each choice, they saw both the payoff obtained from their selection and the payoff they would have received had they selected the forgone option. Note that this full feedback structure reduces the trade-off between exploration and exploitation.

The left hand side of table \ref{tab_tasks} displays the four choice-tasks used. These tasks were selected to examine the behavioral phenomena mentioned above. Specifically, Tasks 1 \& 2 include consequential rare events, and can therefore be used to investigate underweighting of rare events, surprise triggers change (since rare events are by definition surprising) and the wavy recency effect of rare events. Note that in both tasks, choosing as if rare events are underweighted (selecting \textit{B} in Task 1, and \textit{A} in Task 2) implies deviation from maximization of expected payoffs.  Tasks 3 \& 4, that involve correlated outcomes, were included to examine the emergence of the correlation effect: much faster learning to maximize (to select \textit{B} in either task) in Task 3 than in Task 4. 

All decision makers were presented with the following introduction to the experiment: \texttt{The current experiment includes many trials. Your task, in each trial, is to select one of two options, A or B. Each selection will be followed by the presentation of both choices' payoffs. Your payoff for the trial is the payoff of the selection you made}. The human participants then saw two buttons labeled \textit{A} and \textit{B} with text asking for a choice, while the LLMs were asked to respond with their selection and additional instructions to enforce a one-character text response. After choosing, decision makers saw this feedback: \texttt{You selected \{$x$\} and received a payoff of \{$y$\}. Had you selected \{$w$\}, you would have received \{$z$\}}, where $x$, $y$, $w$ and $z$ correspond with the specific trial's decisions and outcomes. Decision makers were then moved to the next trial. To eliminate order effects, we counter-balanced the order of presentation of the options across decision makers.\footnote{Indeed, LLMs, particularly the GPT models, were somewhat biased towards the option presented first, but this pattern is outside of our scope here.}

\subsection{LLM Agents}
Two proprietary large language models were evaluated - OpenAI's GPT-4o mini (2024-07-18 update) and Google's Gemini-1.5 Flash-002, selected for their cost-effectiveness at state-of-the-art level performance. Responses from agents of temperatures $0.2$, $1$ and $2$ were elicited via an API.\footnote{Temperature is defined as the likelihood of a lower-probability output response being selected, with a value of 0 resulting in the highest probability output always being selected. A temperature of $1$ is the default in a range of \texttt{[0, 2]}.} 

The context, the manner in which the agent received their choice history, was considered in one of two paradigms: \textit{chat} or \textit{all}. The \textit{chat} paradigm replicates the chat-bot style conversation, popular on the ChatGPT or Gemini web interfaces, by which the agent has access to the entire history of messages but does not explicitly receive them in a new prompt. This method is most similar to how a human experiences these decision tasks (in most DFE studies, as well as in ours) - the history of choices and outcomes may exist in memory, but it is not explicitly provided. Alternatively, the \textit{all} paradigm explicitly provides the full history of choices and outcomes from the start of the experiment, in each new prompt.\footnote{The additional prompt read: ``\texttt{The following summarizes the past rounds, where `choice' is what you chose, `outcome' is what you received, and `alternative' is what you would have received, had you chosen the other option. The numbers represent which round this occurred in}", and was followed by a dictionary of the form \{trial number, choice, outcome, alternative\} for each trial.}. In this manner, the agent is forced to confront the entire history of outcomes at each decision. We refer to agents of a specific model and context combination as \textit{model-context} (e.g., a GPT-4o mini provided with the entire choice history is referred as \textit{GPT-all}). 

If an agent responded with an answer other than \textit{A} or \textit{B}, it was instructed to retry using the same prompt. If an agent failed to provide a valid response 7 times in a row, the agent was rejected and its choices discarded.
Each decision task was evaluated on 40 agents of each model and context, resulting in 160 agent-experiments per task. Each agent faced only one decision task.  

\subsection{Human Participants}
110 human participants were recruited via Prolific. Each participant encountered exactly one of the four tasks, with each task faced by at least 25 different participants. Participants were paid £0.75 for, on average, 4.6 minutes it took them to complete the 100-trial experiment. In addition, they could have won a  bonus of £0.5. The probability to win the bonus was a linear function of the number of points they accumulated in the experiment. Participants knew that to maximize their chances of winning the bonus, they should earn as many points as possible.

\subsection{Decision Strategies}\label{meth_analysis}
To better understand the strategies employed by the decision makers, their choices were compared to the predicted choices of three theoretical ``players'' employing each a specific decision strategy: The \textit{Recency Player}, choosing the option that led to the best payoff in the previous round only; the \textit{Fictitious Player}, choosing the option with the higher average payoff observed so far in the experiment \cite{brown1951iterative}; and the \textit{WSLS Player}, based on the Win-Stay Lose-Shift strategy \cite{nowak1993win}, in which a player only shifts their current choice in the absence of a positive payoff, even if the alternative would have provided a higher one.  

\section{Results}
We found very little impact of the temperature of the LLM agents, and thus all reported results refer to the default temperature of 1. The five rightmost columns in Table \ref{tab_tasks} summarize the aggregate choice rates of \textit{Option B} (the risky option in Tasks 1 \& 2 and the payoff-maximizing option in Tasks 3 \& 4), for each model-context combination and for the human participants. The results suggest that, on the aggregate, the models and the humans display similar behavioral phenomena (albeit to different extents): Both LLMs and humans exhibit underweighting of rare events (Tasks 1 \& 2), and the correlation effect (Tasks 3 \& 4). In contrast, as we show below, analysis of the sequential dependencies (i.e., of the choice rates contingent on recent events) reveals that LLMs behave very differently than humans on a trial-by-trial basis. 

\subsection{Underweighting of Rare Events}
 All models, and the human participants, chose the risky option (\textit{B}) in Task 1 and the safe option (\textit{A}) in Task 2 more often than not, despite these options having lower expected values than their alternatives. These deviations from maximization imply both LLMs and humans behave as if they underweight the rare events. Notably, in most cases, the choice rates are more extreme for the LLMs than for the humans, indicating, somewhat surprisingly, that the LLMs deviate from payoff maximization even more so than humans. 

\subsection{Correlation Effect}
The aggregate choice rates of both LLMs and humans in Tasks 3 \& 4 are consistent with the correlation effect \cite{correlation-effect}: all decision makers have higher maximization rates with positive (Task 3) than with negative (Task 4) correlations between the options. Specifically, in Task 3, with a positive correlation, Option B never provides a worse payoff than Option A (in each trial, either \textit{B} provides 8 and \textit{A} provides 6, or both options provide 0). All models and the humans indeed learn to maximize in this task, although only the humans and \textit{Gemini-all} have high maximization rates that would match expectations. In Task 4, with a negative correlation, learning to maximize becomes harder for all decision makers. Indeed, the maximizing Option \textit{B} only provides a higher payoff than Option \textit{A} in 50\% of the trials (in each trial, either \textit{B} provides 9 and \textit{A} provides 0, or \textit{B} provides 0 and \textit{A} provides 6), which hinders learning. All models, barring \textit{Gemini-all}, who manages to learn and maximize (as do humans), albeit very slowly, choose \textit{B} approximately 50\% of the time, matching the probabilities of the higher outcome in the task. This may hint that the models rely on the feedback from the most recent trial.  

\subsection{Wavy Recency Effect}\label{wavy_res}
\begin{figure}
    \begin{center}
    \includegraphics[width=0.9\columnwidth]{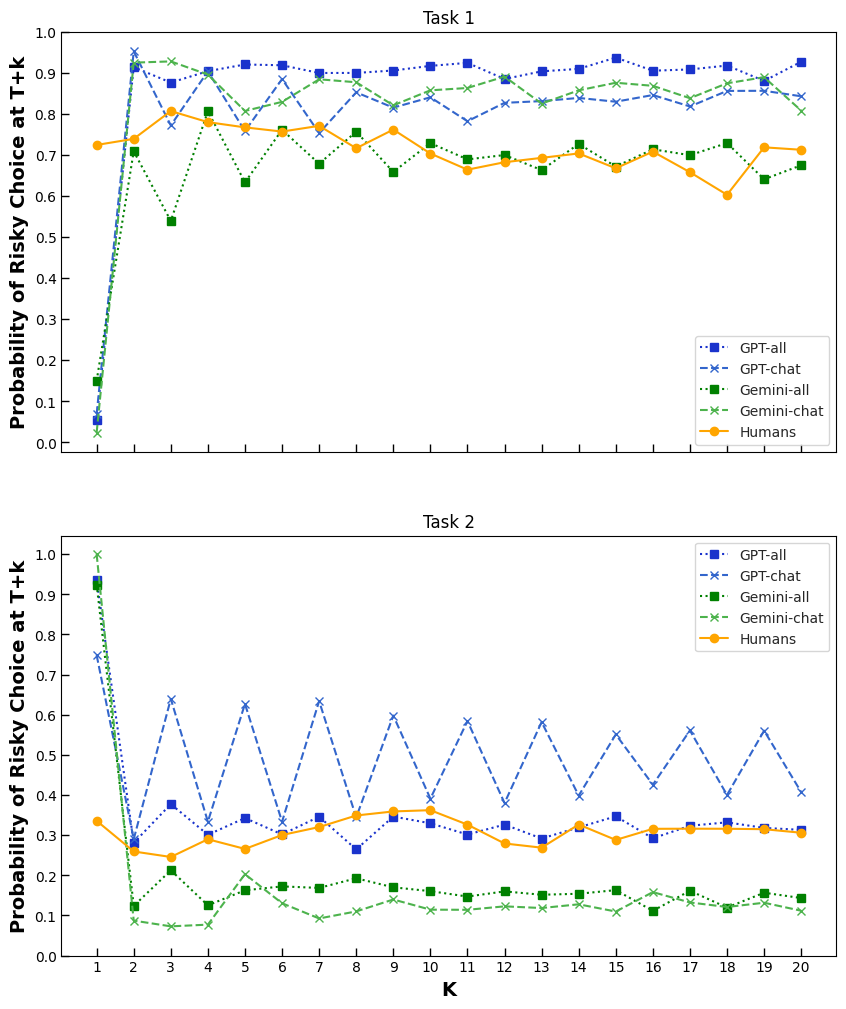}
    \caption{Wavy recency trends, tasks 1 (top) and 2 (bottom)}
    \label{fig_wavy}
    \end{center}
\end{figure}

Figure \ref{fig_wavy} shows the response of the decision makers to seeing the rare events in Tasks 1 and 2. Specifically, it presents the choice rates of the risky option in trials $t+k$ (for $k=1,...20$), contingent on that option generating a rare event in Trial $t$. The humans (orange lines) exhibit a wavy recency pattern consistent with \cite{PLONSKY201718, wavy-recency}. In Task 1 (top panel), the rare event is a large loss (-10), so higher sensitivity to the rare event implies fewer risky choices in this task. In Task 2 (bottom panel), the rare event is a large gain (+10), so higher sensitivity to the rare event implies more risky choices in that task. The figure shows that in both tasks, the lowest human sensitivity to the rare event is observed exactly three trials after it occurs ($k=3$): In Task 1, the risky option is chosen with highest likelihood, and in Task 2, it is selected with the lowest likelihood. Then, in both tasks, the humans behave as if they become increasingly \emph{more} sensitive to the rare event the further back it occurred, with the lowest sensitivity to the rare event long after it happened - a wavy recency effect. 

In sharp contrast to these results, the recency LLM plots do not show clear ``wavy" patterns. Instead, the models all seem to have a strong bias to respond to the rare event in the first trial after it occurs. In Task 1, after a large negative payoff (-10) from the risky option, the models all choose it with very low proportions. In Task 2, after a large positive payoff (+10) from the risky option, the models all choose it with very high proportions. This strong reaction seems to dissipate almost entirely in the next round ($k=2$), where the choice rates all reflect behavior consistent with underweighting (or neglect) of the rare event that was observed merely two trials ago. 

Similar analyses concerning Tasks 3 and 4 (where no wavy recency effect is expected because the tasks do not include rare events; analyses not shown) suggests that the LLMs' maximization rates are highly correlated with the outcomes in the previous trial but not with the outcomes in the trials preceding it. For example, in Task 4, we observe that two trials after the maximizing option \textit{B} provides a better payoff than \textit{A}, the LLMs' maximization rates drop to 50-55\%, reflecting almost complete neglect of the information from before the very recent trial.

Here, we see a deviation between humans and LLMs - while the language models are usually motivated by the most recent outcome, the human decision making process is more complex, and a larger window of experience is considered.  

\subsection{Surprise Triggers Change}
\begin{table}
\caption{Human $B$ choice rates at trial $t+1$ (\%).} 
\label{tab_stc} 
\vskip 0.12in
    \centering\resizebox{\columnwidth}{!}{%
    \begin{tabular}{ccccc}\hline
         Task & $B_t \land V(B_t)>0$ & $B_t \land V(B_t)<0$ & $A_t \land V(B_t)>0$ & $A_t \land V(B_t)<0$ \\
         \hline
         \textbf{1} & 78 & 67 & 48 & 58  \\
         \textbf{2} & 33 & 52 & 31 & 15  \\
         \textbf{3} & 97 & 91 & 60 & 45  \\
         \textbf{4} & 73 & 63 & 70 & 50   \\
         \hline
    \end{tabular}}
    \vspace{0.1cm} % Adds a bit of space between the table and the note
\footnotesize
\textit{Note:} $X_t, (X \in \{A,B\})$ implies choice of Option $X$ in trial $t$. $V(B_t)$ is the payoff provided by Option $B$ in trial $t$. Thus, e.g., column $A_t \land V(B_t)>0$ gives the choice rate of $B$ immediately after trials where Option $A$ was selected but Option $B$ provided a gain.
\end{table}

Table \ref{tab_stc} displays the \textit{B} choice rates of (only) the human participants following various surprising and unsurprising events, and contingent on the participant's most recent choice.
The results suggest that after observing a surprising (i.e. rare) event, humans become more likely to switch their choice from the previous trial, even if the feedback in the previous trial was reinforcing of their recent choice. To see this, consider Task 1. Relative to trials that follow a non-surprising +1 gain, in trials that followed a surprising -10 loss from Option $B$, humans were more likely both to switch away from $B$ if they just chose it (33\% vs. 22\%) and  \emph{switch into $B$} after they chose $A$ (58\% vs. 48\%). That is, more switches to \textit{B} occur even though $B$ just provided a very poor outcome. Similarly, in Task 2, humans were more likely to switch from $A$ to $B$ after seeing $B$ providing a +10 surprising event than after seeing the -1 non-surprising one (31\% vs.15\%), and also more likely to \emph{switch away from $B$} after it provided them a surprising +10 than after it provided them a non-surprising -1 (67\% vs. 48\%). Thus, humans exhibit a ``surprise triggers change" effect. 

Notably, none of the LLM models exhibited this odd counter-reinforcement behavior (analyses not shown), instead responding in alignment with the best observed previous outcome, reflective of their high recency bias.

In Tasks 3 and 4 when nothing is surprising, we observe that both humans and LLM agents choose $B$ more after it provides a gain than after it provides a loss, regardless of their previous choice. 

\subsection{Choice Strategies}
\begin{table}
\caption{Proportion of strategy matching choices (\%)} 
\label{tab_matching} 
\vskip 0.12in
\centering\resizebox{0.95\columnwidth}{!}{%
\begin{tabular}{ccccccc}
         \hline
         Task & Player & Gemini-all & Gemini-chat & GPT-all & GPT-chat & Humans\\
         \hline
         \multirow{3}{2em}{1} & Recency & 81 & 94 & 98 & 91 & 67 \\ 
          & WSLS & 81 & 93 & 98 & 91 & 68 \\
          & Fictitious & 56 & 52 & 56 & 43 & 46 \\\hline
         \multirow{3}{2em}{2} & Recency & 93 & 96 & 76 & 56 & 69 \\ 
          & WSLS & 28 & 27 & 44 & 92 & 22  \\
          & Fictitious & 52 & 53 & 52 & 48 & 59\\\hline
         \multirow{3}{2em}{3} & Recency & 75 & 76 & 66 & 63 & 73 \\ 
          & WSLS & 52 & 83 & 55 & 69 & 52 \\
          & Fictitious & 96 & 67 & 77 & 59 & 91 \\\hline
         \multirow{3}{2em}{4} & Recency & 88 & 100 & 87 & 100 & 58 \\ 
          & WSLS & 88 & 100 & 87 & 100 & 59 \\
          & Fictitious & 63 & 54 & 57 & 55 & 65\\\hline
          \multirow{3}{2em}{Mean} & Recency & 84 & 91 & 82 & 78 & 67 \\ 
          & WSLS & 62 & 76 & 71 & 88 & 50 \\
          & Fictitious & 67 & 57 & 61 & 51 & 65\\
         \hline
    \end{tabular}}

\end{table}

Table \ref{tab_matching} details the proportion of identical choices between each group of decision makers, and the three strategies of play described in the \nameref{meth_analysis} section - Recency Player, Win-Stay Lose-Shift Player, and Fictitious Player, as well as the mean of those matching choices across all tasks. No strategy was consistently played across the board by any model or by humans, indicating that choices were noisier than is assumed by these deterministic strategies. Nonetheless, the language models display a recency bias, sometimes extremely so, on average matching the recency player's choices in 78\%-91\% of the decisions. This recency bias is much stronger than that exhibited by humans.  

\subsection{Learning Trends}
\begin{figure}
    \begin{center}
    \includegraphics[width=\linewidth]{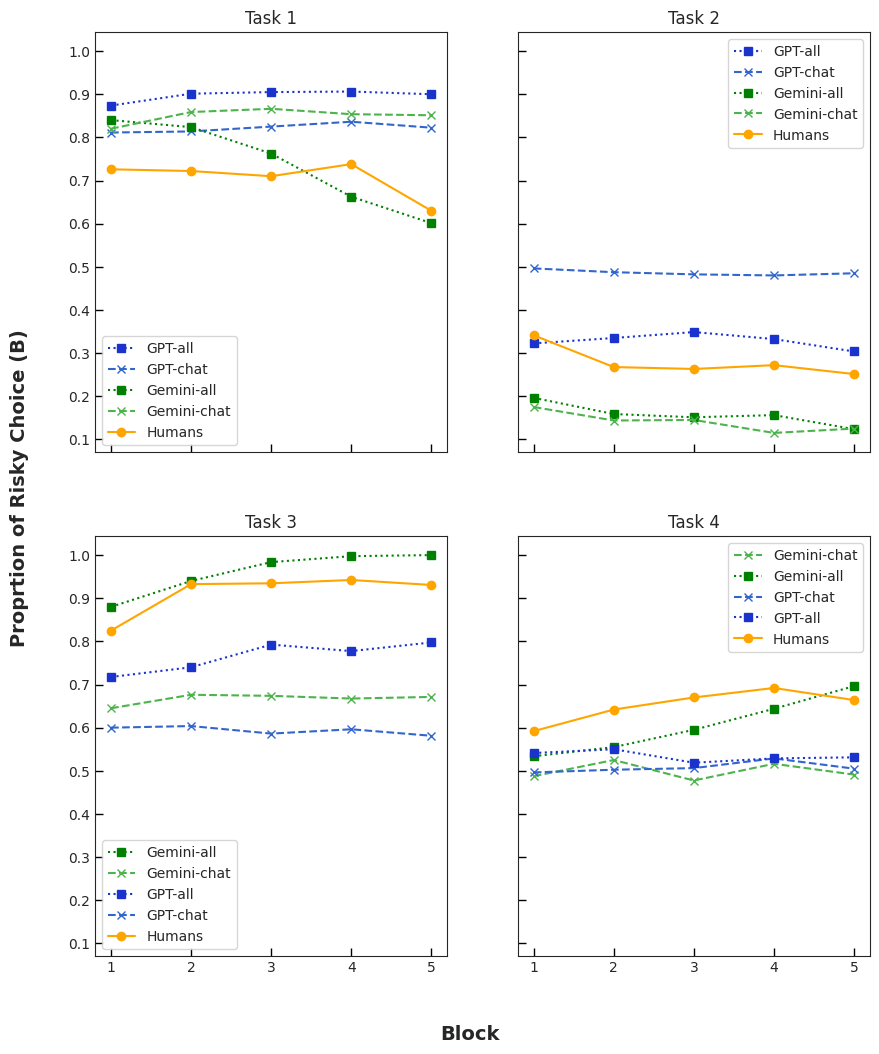}
    \caption{Choice rate of maximizing option over experiment blocks}
    \label{fig-learn}
    \end{center}
\end{figure}

Figure \ref{fig-learn} displays the \textit{B} choice rates of over the duration of the 100 rounds for each task, divided into 5 blocks of 20 choices.
In most cases, the maximization rates do not significantly change over the course of the experiment. The main exceptions are the humans and \textit{Gemini-al}l in Tasks 3 and 4. Notably, the learning patterns differ in these cases as well: While humans learn mostly early on in the experiment,\textit{Gemini-all}'s learning is more steady and continuous. In the other tasks, learning is slow. For example, in Task 1, the maximization rates of both humans and \textit{Gemini-All} are higher at the end than in the beginning of the experiment, but both still deviate from maximization (i.e., behave as if underweighting of rare events) in the last block. In addition, comparison of the \textit{all} and \textit{chat} context paradigms suggests somewhat better learning of the LLM agents under \textit{all} than under \textit{chat}, but the differences are inconsistent and not large.

\section{Discussion}
It is evident that, on average, LLMs exhibit similar behavioral characteristics to humans in DFE tasks, but to a different degree and importantly, for very different reasons. Both humans and LLMs behave as if they underweight rare events, but LLMs do so to a greater degree. Similarly, the correlation effect is observed in both humans and LLMs, but the rate of learning in both cases is much faster in humans (with the notable exception of \textit{Gemini-all}). Here, however, is where the parallels end. Other decision phenomena that are robustly observed in humans, are not present in LLMs at all - the wavy recency effect and surprise triggers change.

It stands to reason that the language models are driven by an extreme recency bias, rarely considering the payoffs of either options that occurred more than a few trials ago. It is striking to note that this extreme recency degree is still present in the event in which the model receives the entire choice and outcome history in every prompt, (although perhaps to a lesser degree, evidenced by \textit{Gemini-All}'s successful maximization in Tasks 3 and 4). In this case, it would have been straightforward to calculate the option that has generated a better payoff on average (i.e., to play as a fictitious player), but the models do not do this.

This recency preference can explain the differences we observe between humans and LLMs. The language models underweight rare events to a more extreme degree because they mainly consider the most recent of outcomes, and as a result, would choose options in proportion to how often they ``win". In our tasks, this implies 90\% choice of the option that includes consequential rare events. The wavy recency effect and surprise triggers change are also absent, as no waviness pattern can emerge when the sensitivity to the last result is always the highest, and since surprise triggers change implies deviation from the bias towards the most recently successful outcomes. Note, however, that a lack of surprise triggers change effect in LLMs may be expected given that surprise is triggered by one's expectation of the event, and language models do not have expectations about outcomes, they simply respond to prompts.

Both the model used in an LLM agent, and the context it is provided, play a significant role in the choices the agent makes. The explanation for the difference between questioning paradigms is intuitive - the strong recency bias is slightly mitigated by passing the entire choice history to the agent in the prompt, but not in a manner in which all rounds are considered.  

Importantly, the wavy recency effect of rare events in humans is explained by a sophisticated attempt to identify patterns in the sequences of observed outcomes. When these patterns do not exist, this attempt would lead both to the wavy recency effect and to underweighting of rare events \cite{wavy-recency}. With that in mind, it is notable that LLMs fail to exhibit the same effect, potentially suggesting humans behave more in a more sophisticated manner in DFE experiments than LLMs do. One way to test this in future work, would be to compare humans and LLMs in DFE tasks where sequential patterns in the outcomes (e.g., alternation) truly exist. The hypothesis raised above suggests that humans but not LLMs would pick up on these patterns.

A possible explanation for our findings may be the nature of an Autoregressive Language Model, whose task is outputting the words with the highest probability of being the next word in the sequence\footnote{Randomness, alignment, censorship and fine-tuning also playing a role, but the fundamental task remains the same.}. This would intuitively result in a recency bias - in a regular conversation, a fact mentioned 15 sentences ago may provide context to the discussion, but would not necessarily be considered when evaluating the most likely word to follow the most recently mentioned one. Therefore, a model may naturally ignore past outcomes, as it does not consider them as important for the task at hand.

As the field of language models advances rapidly, a continuous reevaluation of these findings is warranted, alongside an expansion of the scope of this research, both in terms of the models used and the decision scenarios tested. Future works can ask the models to portray a specific persona, or predict the choices a human would make in the given scenario, which are established means of assessment (e.g \citeNP{aher2023, new-paper}) that can add another layer of insight into the behaviors and choice patterns of LLM agents. Specifically, these methods have not yet been tested on decision from experience tasks or many-shot problems. Finally, fine-tuning LLMs to experience-related decision tasks is likely to display patterns more robustly comparable to the choices of humans.

\section{Conclusion}
LLMs seem, on the surface, very human-like. An analysis of their aggregate choices in Decisions From Experience tasks would initially confirm this. However, we have shown how they exhibit the similar tendencies for very different reasons than humans do - a strong recency bias. That is, our nuanced assessment of their choice rates contingent on recent events reveals that LLMs behave very differently than humans. LLMs have enormous potential to replicate, simulate and aid in the research of human behavior, but require careful investigation and analysis to truly understand how robust the similarities are.

\bibliographystyle{apacite}

\setlength{\bibleftmargin}{.125in}
\setlength{\bibindent}{-\bibleftmargin}

\bibliography{CogSci_Template}
\end{document}